\documentclass[10pt,twocolumn,letterpaper]{article}

\usepackage[pagenumbers]{iccv} 

\usepackage{multirow}
\usepackage{algorithm}
\usepackage{algorithmic}
\usepackage{stfloats}
\usepackage{listings}
\usepackage{bbm}
\usepackage{pifont}
\newcommand{\xmark}{\ding{55}}
\definecolor{iccvblue}{rgb}{0.21,0.49,0.74}
\usepackage[pagebackref,breaklinks,colorlinks,allcolors=iccvblue]{hyperref}

\def\confName{ICCV}
\def\confYear{2025}

\title{One Object, Multiple Lies: A Benchmark for Cross-task Adversarial Attack on Unified Vision-Language Models}

\author{Jiale Zhao$^{1}$ \quad Xinyang Jiang$^{2}$\thanks{\quad Corresponding author} \quad Junyao Gao$^{1}$ \quad Yuhao Xue$^{1}$  \quad Cairong Zhao$^{1*}$ \\
$^{1}$Tongji University\quad
$^{2}$Microsoft Research Asia \\
\tt\small\{2410917, junyaogao, 2432200, zhaocairong\}@tongji.edu.cn,\\ 
\tt\small xinyangjiang@microsoft.com
}

\begin{document}
\maketitle

\begin{abstract}
Unified vision-language models (VLMs) have recently shown remarkable progress, enabling a single model to flexibly address diverse tasks through different instructions within a shared computational architecture. This instruction-based control mechanism creates unique security challenges, as adversarial inputs must remain effective across multiple task instructions that may be unpredictably applied to process the same malicious content. 
In this paper, we introduce CrossVLAD, a new benchmark dataset carefully curated from MSCOCO with GPT-4-assisted annotations for systematically evaluating cross-task adversarial attacks on unified VLMs. CrossVLAD centers on the object-change objective—consistently manipulating a target object's classification across four downstream tasks—and proposes a novel success rate metric that measures simultaneous misclassification across all tasks, providing a rigorous evaluation of adversarial transferability.
To tackle this challenge, we present CRAFT (Cross-task Region-based Attack Framework with Token-alignment), an efficient region-centric attack method.  Extensive experiments on Florence-2 and other popular unified VLMs demonstrate that our method outperforms existing approaches in both overall cross-task attack performance and targeted object-change success rates, highlighting its effectiveness in adversarially influencing unified VLMs across diverse tasks. Code is available at \href{https://github.com/Gwill-Z/CRAFT}{https://github.com/Gwill-Z/CRAFT}.
\end{abstract}    
\section{Introduction}
\label{sec:intro}



Recent advances in unified vision-language models (VLMs) have demonstrated remarkable capabilities in handling multiple downstream tasks through different instructions~\cite{xiao2024florence, wang2022ofa, lu2022unified, lu2024unified, shukor2023unival, li2023uni, zeng2023x, gao2024styleshot, gao2025faceshot}. These unified VLMs adopt a sequence-to-sequence framework to model diverse tasks using shared architecture and parameters, controlled by different task-specific instructions. Their versatility has attracted significant research interest and shown promising potential in applications from autonomous driving~\cite{huang2024vlm} to smart city systems~\cite{ci2023unihcp, he2024instruct, jin2024you}. However, as these unified VLMs continue to evolve, understanding their security vulnerabilities becomes increasingly critical for ensuring reliability in diverse operational scenarios, particularly their susceptibility to adversarial attacks.



When considering adversarial attacks against unified VLMs, their instruction-based nature creates a unique security challenge: adversarial inputs must remain effective regardless of which task instructions are later applied, as attackers typically cannot predict which instructions will process their malicious inputs. This necessitates cross-task adversarial attacks, where a single adversarial example must successfully manipulate model outputs across multiple different tasks simultaneously. 
As illustrated in Figure~\ref{benchmark}, an adversarial example demonstrating strong cross-task transferability would successfully deceive the model across all four representative tasks—causing incorrect predictions in captioning, detection, region classification, and object localization—despite being subjected to different task instructions.

Recent efforts to attack VLMs have explored various techniques, from cross-modal guidance and optimal transport optimization~\cite{lu2023set, han2023ot} to intermediate feature attacks~\cite{hu2024firm}, primarily focusing on enhancing adversarial transferability across models. While considerable research examines model transferability—the ability of adversarial examples to generalize across different models performing the same task—the critical dimension of cross-task transferability within unified VLMs remains largely unexplored.
Although some studies have explored attacks beyond single tasks, they can be categorized into two distinct approaches that differ from our work. The first category—multi-task attacks~\cite{zhe2024adversarial, zeng2024cross, feng2024enhancing, wang2021psat, gao2023similarity}—generates separate adversarial examples optimized for different tasks independently, rather than creating a single adversarial example capable of attacking multiple tasks simultaneously as in our cross-task approach. The second category includes works that attempt cross-task attacks similar to ours~\cite{cui2024robustness, xu2024highly, lu2023set, hu2024firm}, but their evaluation methodology remains fundamentally limited: they assess attack performance separately for each task, reporting individual success rates rather than measuring whether a single adversarial example successfully fools multiple tasks simultaneously. This evaluation gap is critical for unified VLMs with instruction-controlled architectures, as it fails to capture the more strict scenario where the same perturbed image is required to consistently manipulate model outputs across all potential task instructions it might encounter. 

To address this critical research gap, we introduce CrossVLAD, a comprehensive benchmark specifically designed to evaluate cross-task transferability of adversarial attacks on unified VLMs. Built upon popular vision-language datasets MSCOCO with GPT-4-assisted annotations, CrossVLAD encompasses four representative tasks: image captioning, object detection, region category recognition, and object location prediction.These four tasks were strategically selected to cover the core functional capabilities of unified VLMs, representing different combinations of textual and spatial reasoning required across typical vision-language applications.
Central to CrossVLAD is a novel cross-task object manipulation attack scenario, where adversarial examples must consistently manipulate a target object's identity across all tasks simultaneously. To quantitatively measure cross-task transferability, we propose two new metrics: CTSR-4 (Cross-Task Success Rate-4) for the proportion of samples that successfully fool all four tasks, and CTSR-3 for samples that compromise at least three tasks.
Based on this benchmark, we present {\bf CRAFT} (Cross-task Region-based Attack Framework with Token-alignment), an efficient attack method specifically designed for unified VLMs. CRAFT employs a region-centric approach with cross-task feature alignment to generate adversarial examples that can successfully attack multiple vision-language tasks simultaneously.


Our main contributions can be summarized as follows:
\begin{itemize}
    \item We present a systematic study of cross-task adversarial attacks on unified VLMs, introducing CRAFT, a novel method that effectively exploits the unified representation space to perform object-change attacks—consistently manipulating a target object's identity across diverse vision-language tasks.
    
    \item We propose CrossVLAD, the first benchmark for evaluating cross-task adversarial transferability in unified VLMs, featuring carefully curated datasets and novel evaluation metrics (CTSR-4 and CTSR-3) that provide comprehensive measures of attack success across multiple tasks. 

    \item Through extensive experiments on state-of-the-art unified VLMs including Florence-2, OFA, and UnifiedIO-2, we demonstrate that CRAFT significantly outperforms existing methods in terms of both overall cross-task attack performance and targeted object manipulation success rates.
\end{itemize}

\section{Related Works}
\subsection{Unified VLMs}
Recent years have witnessed significant advancements in unified vision-language models (VLMs), which map multi-modal data into a unified representation space for diverse tasks such as image captioning, object detection, and visual grounding \cite{xiao2024florence, wang2022ofa, lu2022unified, lu2024unified, shukor2023unival, li2023uni}. Florence-2 \cite{xiao2024florence} leverages large-scale pretraining to enhance multi-task capabilities, while OFA \cite{wang2022ofa} introduces sequence-to-sequence learning without task-specific layers. UNIFIED-IO 2\cite{lu2024unified} converts heterogeneous inputs into token sequences, and X2-VLM \cite{zeng2023x} optimizes multi-grained vision-language alignments. Domain-specific applications have emerged in human-centric tasks \cite{ci2023unihcp} and autonomous driving \cite{huang2024vlm}, demonstrating the versatility of unified VLMs while introducing new security challenges, particularly regarding adversarial vulnerabilities.

\subsection{Transferability of Adversarial Attacks}
Adversarial attacks on Vision-Language Models exhibit concerning transferability properties that amplify their security implications \cite{cui2024robustness, liu2024survey, schlarmann2023adversarial}. Cross-model transferability has been demonstrated across various VLMs, including sophisticated models like Bard and GPT-4V \cite{dong2023robust, tu2023many}. Studies show that adversarial examples crafted for one model can successfully compromise others despite architectural differences \cite{wang2024transferable, zhao2023evaluating, bailey2023image}. Attack methods such as cross-prompt manipulations \cite{luo2024image} has proven effective across different VLM implementations, highlighting a systemic vulnerability in the vision-language domain.

Beyond cross-model transfer, research has increasingly focused on cross-task transferability. Existing approaches generally fall into two categories: multi-task attacks \cite{zhe2024adversarial, zeng2024cross, feng2024enhancing, wang2021psat, cheng2024attentional, lv2023ct, lu2024time} that generate separate adversarial examples optimized for different tasks independently, and cross-task attacks \cite{cui2024robustness, xu2024highly, lu2023set, hu2024firm} that attempt to create examples transferable across tasks. Attention manipulation techniques have emerged as particularly effective in this domain, with methods altering model focus across different tasks. Other approaches enhance transferability through diversified perturbations \cite{zhang2022boosting} and exploiting visual relations \cite{ma2023boosting}.

Current research methodology inadequately evaluates unified VLMs' security by measuring attack success rates on individual tasks rather than assessing whether single adversarial examples can simultaneously compromise multiple functionalities. This siloed approach fails to capture how unified representations in unified VLMs create systemic vulnerabilities to coordinated attacks—a critical security gap our work addresses.
\section{CrossVLAD Benchmark}
We introduce CrossVLAD, the first benchmark designed to evaluate cross-task adversarial attacks on unified vision-language models. Our benchmark addresses a sophisticated threat model where adversarial perturbations simultaneously manipulate an object's identity across diverse tasks. CrossVLAD encompasses four representative tasks, introduces metrics for measuring cross-task attack success, and includes 3,000 carefully curated samples. The following subsections detail our task definitions, dataset construction process, and evaluation methodology.

\subsection{Task Definition and Attack Objective}

\subsubsection{Task Definition}
CrossVLAD evaluates cross-task adversarial transferability through four representative vision-language tasks: (1) \textbf{Image Captioning} ($T_{cap}$): generating textual descriptions of entire images; (2) \textbf{Object Detection} ($T_{det}$): identifying and localizing all objects in images; (3) \textbf{Region Categorization} ($T_{reg}$): classifying specific image regions; and (4) \textbf{Object Localization} ($T_{loc}$): finding regions corresponding to specified object categories. These tasks form complementary pairs—captioning and detection operate globally across the entire image, while region categorization and object localization focus on specific regions—allowing comprehensive assessment of adversarial effects across both semantic understanding and spatial reasoning processes.

\subsubsection{Object-change Attack Objective}
We focus on the object-change attack scenario against unified VLMs. Given a model $\mathcal{M}$ capable of performing multiple tasks $\mathcal{T} = \{T_{cap}, T_{det}, T_{reg}, T_{loc}\}$, our objective is to generate an adversarial example $I_{adv} = I + \delta$ (where $\|\delta\|_\infty \leq \epsilon$) such that a source object of category $c_s$ is consistently misidentified as target category $c_t$ across all tasks. Formally, we aim to manipulate the model outputs $\mathcal{M}(I_{adv}, T_i)$ for each task $T_i \in \mathcal{T}$ to reflect the target category $c_t$ instead of the true source category $c_s$. In summary, our attack objective can be formulated as:

\begin{align}
\underset{\delta: \|\delta\|_\infty \leq \epsilon}{\text{maximize}} \prod_{i \in \{cap, det, reg, loc\}} S_i(I + \delta, c_s, c_t)
\end{align}
where $S_i$ represents the attack success function for task $T_i$ as defined in Section~\ref{single_eval}.

\subsection{Dataset Construction}
We constructed CrossVLAD from the MSCOCO train2017 dataset~\cite{lin2015microsoftcococommonobjects} to evaluate cross-task adversarial attacks. The benchmark contains 3,000 carefully selected images with 79 change-pairs across 10 semantic categories (Vehicle, Outdoor, Animal, Accessory, Sports, Kitchen, Food, Furniture, Electronic, and Appliance). These pairs define semantically reasonable transformations (e.g., "bicycle" to "motorcycle", "cat" to "dog") for diverse attack scenarios.

Our selection process implemented several key criteria: (1) object size constraints (10\%-50\% of image area); (2) limited object instances per image  with category uniqueness; (3) caption verification (object mentioned of 5 captions); and (4) exclusion of images containing potential target categories. For annotation, we preserved original MSCOCO annotations and employed GPT-4 to generate ground-truth captions for the target object category. Each generated caption was verified to explicitly mention the target category while excluding the source category. Detailed construction procedures, statistics, and the complete list of change-pairs are provided in Appendix~\ref{appendix:dataset}.

\subsection{Evaluation Metrics}

\subsubsection{Single-task Evaluation Criteria}
\label{single_eval}
For each task $T_i \in \mathcal{T}$, we define a task-specific success function $S_i(I_{adv}, c_s, c_t) \in \{0, 1\}$ that evaluates whether the attack successfully manipulates the model output:

\begin{itemize}
    \item \textbf{Image Captioning} ($S_{cap}$): Let $C_{adv}$ be the generated caption. Success requires that the caption includes the target category while excluding the source category. Formally:
    \begin{align}
    S_{cap}(I_{adv}, c_s, c_t) = \mathbbm{1}[c_t \in C_{adv} \wedge c_s \notin C_{adv}]
    \end{align}
    
    \item \textbf{Object Detection} ($S_{det}$): Let $\mathcal{D}(I_{adv})$ be the set of detected objects and $b_s$ be the source object's box. The attack is considered successful if an object of the target category is detected at the source location:
    \begin{align}
    S_{det}(I_{adv}, c_s, c_t) = \mathbbm{1}[&\exists (b_i, l_i) \in \mathcal{D}(I_{adv}) : \nonumber \\
    &\text{IoU}(b_i, b_s) > \theta_{box} \wedge l_i = c_t 
    \end{align}
    
    \item \textbf{Region Categorization} ($S_{reg}$): Given source box $b_s$, let $l_{reg}$ be the predicted category. A successful attack occurs when the model misclassifies the specified region as the target category:
    \begin{align}
    S_{reg}(I_{adv}, c_s, c_t) = \mathbbm{1}[l_{reg} = c_t]
    \end{align}
    
    \item \textbf{Object Localization} ($S_{loc}$): Given target category $c_t$, let $b_{loc}$ be the predicted box. We define success as the model returning a bounding box for the target category that substantially overlaps with the original source object's position:
    \begin{align}
    S_{loc}(I_{adv}, c_s, c_t) = \mathbbm{1}[\text{IoU}(b_{loc}, b_s) > \theta_{loc}]
    \end{align}
\end{itemize}

\subsubsection{Cross-task Evaluation Metrics}
To evaluate cross-task transferability, we define two primary metrics:

\begin{itemize}
    \item \textbf{CTSR-4} (Cross-Task Success Rate-4): The percentage of examples that fool all four tasks:
    \begin{align}
    \text{CTSR-4} = \frac{1}{N}\sum_{j=1}^{N} \prod_{i \in \mathcal{I}} S_i(I_{adv}^j, c_s^j, c_t^j)
    \end{align}
    where $\mathcal{I} = \{cap, det, reg, loc\}$ and $N$ is the total number of test samples.
    
    \item \textbf{CTSR-3} (Cross-Task Success Rate-3): The percentage of examples that fool at least three tasks:
    \begin{align}
    \text{CTSR-3} = \frac{1}{N}\sum_{j=1}^{N} \mathbbm{1}\left[\sum_{i \in \mathcal{I}} S_i(I_{adv}^j, c_s^j, c_t^j) \geq 3\right]
    \end{align}
\end{itemize}

These metrics provide a comprehensive assessment of an attack's ability to coherently manipulate the model across multiple tasks.
\section{Method}
\subsection{Overview}
Based on our CrossVLAD benchmark and object-change attack objective, we propose CRAFT (Cross-task Region-based Attack Framework with Token-alignment), a method that exploits the unified representation space of unified VLMs to generate adversarial examples effective across multiple vision-language tasks. As shown in Figure~\ref{benchmark}, given an input image $I$ with a source object of category $c_s$ and a target category $c_t$, CRAFT first identifies the token regions corresponding to the source object, then extracts text embeddings for relevant categories and optimizes the perturbation $\delta$ using a contrastive alignment loss that simultaneously pushes the object's visual representation toward the target category and away from the source and other negative categories. The resulting adversarial example $I_{adv} = I + \delta$ ($\|\delta\|_\infty \leq \epsilon$) is designed to manipulate the model's outputs across all four tasks, leading to consistent misidentification of the source object as the target category. The complete algorithm procedure is provided in Appendix~\ref{appendix:algorithm}.

\subsection{Region Token Localization}

Precisely localizing the tokens corresponding to the target object region is crucial for effective cross-task adversarial attacks. Unlike full-image attacks that introduce noise across the entire image, our region-based approach concentrates perturbations on the most semantically relevant areas, offering several advantages: (1) it minimizes visual artifacts in non-target regions, (2) increases attack efficiency by reducing the perturbation search space, and (3) enhances cross-task effectiveness by focusing on the specific tokens that influence object identity across all tasks.

To implement region localization, we first need to map the object's bounding box coordinates in pixel space to token indices in the model's feature space. Given a source object with bounding box $b_s = (x_1, y_1, x_2, y_2)$ in the original image $I$ of size $H \times W$, we compute the corresponding token indices after the image preprocessing and embedding operations. Most unified VLMs first resize the input image to a standard size (e.g., $H' \times W'$) and then divide it into a grid of non-overlapping patches of size $P \times P$, resulting in a sequence of $\frac{H' \times W'}{P^2}$ tokens.

We calculate the token region $\mathcal{R}$ by projecting the original bounding box coordinates to the token grid:
\begin{align}
i_{min} &= \lfloor \frac{x_1 \cdot W'}{W \cdot P} \rfloor, \quad i_{max} = \lceil \frac{x_2 \cdot W'}{W \cdot P} \rceil \\
j_{min} &= \lfloor \frac{y_1 \cdot H'}{H \cdot P} \rfloor, \quad j_{max} = \lceil \frac{y_2 \cdot H'}{H \cdot P} \rceil
\end{align}

The token indices corresponding to the region are then obtained by converting the 2D grid positions to 1D indices in the token sequence. This precise localization ensures that our adversarial perturbations directly target the tokens that most strongly influence the model's representation of the object across different tasks.

\subsection{Cross-modal Feature Alignment}

The core of our attack strategy lies in manipulating the feature representations of the localized region tokens to alter the perceived object identity across multiple tasks. We achieve this through cross-modal feature alignment, leveraging the unified representation space that vision-language models use for both visual and textual information.

Given the localized region tokens $\mathcal{R}$ corresponding to the source object, we extract their feature representations $F_R$ from the output of the image encoder. Simultaneously, we obtain the text embeddings for the target category $c_t$ as positive embeddings $E_{pos}$ and the embeddings for the source category $c_s$ along with other categories as negative embeddings $E_{neg}$ using the model's text encoder.

To align the region features with the target category while pushing them away from the source and other categories, we employ a contrastive loss function:
\begin{align}
\mathcal{L}_{contrast} = \max(0, \text{sim}(F_R, E_{neg}) \nonumber \\ - \text{sim}(F_R, E_{pos}) + \tau)
\end{align}
where $\text{sim}(\cdot, \cdot)$ represents the cosine similarity between feature vectors, and $\tau$ is a margin hyperparameter. This loss encourages the region features to have higher similarity with the target category embedding and lower similarity with the negative category embeddings.

To generate the adversarial example, we use Projected Gradient Descent (PGD)~\cite{madry2017towards} with the computed gradients of the contrastive loss with respect to the input image. At each iteration, we update the adversarial image using:
\begin{align}
I_{adv}^{t+1} = \text{Clip}(I_{adv}^{t} + \alpha \cdot \text{sign}(\nabla_{I_{adv}^t}\mathcal{L}), I-\epsilon, I+\epsilon)
\end{align}
where $\alpha$ is the step size and $\epsilon$ is the perturbation budget.

This approach is particularly effective for cross-task attacks because it directly manipulates the unified feature representations that underlie all tasks in unified VLMs. By aligning the visual features of the source object with the textual features of the target category, we induce the model to consistently misclassify the object across all downstream tasks. 
\begin{figure*}[ht]
\centering
\includegraphics[width=\linewidth]{fig/demo_5.pdf}
\caption{Qualitative examples of CRAFT attack on Florence-2 model across four vision tasks: object detection, object localization, image captioning, and region categorization, with $\varepsilon=16/255$.}
\label{demo}
\end{figure*}

\section{Experiments}


\subsection{Experimental Setup}

\textbf{Attacked Models} \quad We evaluate our method on four representative unified vision-language models: Florence-2-Large \cite{xiao2024florence}, UnifiedIO-2-Large \cite{lu2024unified} and OFA-Large \cite{wang2022ofa}. All models are used with their original pre-trained weights without any task-specific fine-tuning to evaluate the impact of adversarial attacks on pre-trained unified models.

\noindent
\textbf{Compared Methods} \quad Following \cite{zhang2024anyattack}, we compare our approach with several representative attack methods. Since there are currently no existing methods specifically designed for cross-task targeted attacks on unified VLMs, we select state-of-the-art targeted attack methods that can be adapted to our object-change objective. The baseline is the original clean images with no attack. We utilize the text description attack from Attack-Bard \cite{dong2023robust}, which minimizes the training loss between model outputs and target text. Additionally, we modify the adversarial attack method from Mix.Attack \cite{tu2023many} for our targeted scenario. MF-ii \cite{zhao2023evaluating} leverages GPT-4-generated target descriptions to create target images, guiding adversarial samples toward their feature space. Lastly, MF-it \cite{zhao2023evaluating} directly uses target descriptions to optimize the feature representations of adversarial samples. All these methods can potentially manipulate object identity but were not originally designed to maintain consistency across multiple tasks. More implementation details of the compared methods are provided in the appendix~\ref{exp_detail}.

\noindent
\textbf{Implementation Details} \quad In our experiments, we adopt the $\ell_\infty$ norm constraint with $\epsilon = 16/255$ for perturbation, employing the PGD algorithm with 100 iterations and a step size of $\alpha = 4/255$. The margin parameter $\tau$ in the contrastive loss function is set to 0.9. For region localization, we compute region tokens that fully contain the target object. We evaluate attack performance using CTSR-4 and CTSR-3 metrics, which measure the proportion of samples that successfully attack all four tasks or at least three tasks simultaneously. We also report Average Success Rate (ASR), the average of individual success rates across tasks. For bounding box evaluations, we set IoU thresholds $\theta_{box}$ and $\theta_{loc}$ to 0.6. All experiments are conducted using two NVIDIA RTX 3090 GPUs.



\subsection{Benchmarking on Cross-task Scenarios}

Table~\ref{main} presents the performance comparison between CRAFT and existing targeted attack approaches adapted to our cross-task setting. Our results demonstrate that CRAFT consistently outperforms all compared methods across all evaluated models, achieving the highest CTSR-4 and CTSR-3 scores on Florence-2, UnifiedIO-2 and OFA.
 The performance gap is particularly evident in cross-task metrics. While some baselines achieve reasonable success on individual tasks, their low CTSR-4 scores confirm this success does not translate to effective cross-task manipulation. This validates our hypothesis that traditional attacks fail to exploit the unified representation spaces of these VLMs. 
 Figure~\ref{demo} shows qualitative examples where CRAFT successfully manipulates model predictions across all four tasks while introducing only imperceptible perturbations to the original images, demonstrating its effectiveness in creating visually subtle yet semantically consistent adversarial examples.More successes and failures can be found in the appendix~\ref{exp_detail}.

\begin{table*}[ht]
\small
\centering
\begin{tabular}{ccccccccc}
\hline
\multirow{2}{*}{Models}      & \multirow{2}{*}{Method} & \multicolumn{4}{c}{Tasks}                                        & \multicolumn{3}{c}{Evaluate Metrics}            \\ \cline{3-9} 
                             &                         & IC             & OD            & RC             & OL             & avg            & CTSR-4         & CTSR-3        \\ \hline
\multirow{6}{*}{OFA \cite{wang2022ofa}}         & no attack               &    0.005            &     0.007       &       0.003    &   0.026         &     0.01          &      0    &  0.001       \\
                             & Mix.Attack \cite{tu2023many}              &      0.2          &      0.101         &         0.194       &     0.199      &    0.174   &   0.043      &   0.077    \\
                             & MF-it \cite{zhao2023evaluating}            &   0.133         & 0.102        & 0.129     &  0.262        &  0.157           &      0.027    &       0.086        \\
                             & MF-ii \cite{zhao2023evaluating}             &  0.176           & 0.15              &   0.154         &   0.295         &    0.194       &  0.067   &   0.098      \\
                             & Attack-Bard \cite{dong2023robust}      &    \textbf{0.508}   &      0.211         &         0.184       &     0.101      &    0.251   &   0.087      &   0.106           \\
                             & CRAFT(ours)             &    0.394            &     \textbf{0.319 }    &  \textbf{0.435}      & \textbf{0.426}      & \textbf{0.394}      & \textbf{0.286}      & \textbf{0.429}     \\ \hline
\multirow{6}{*}{UnifiedIO-2 \cite{lu2024unified}} & no attack               & 0.002          & 0.002         & 0              & 0.006          & 0.003          & 0              &  0             \\
                             & Mix.Attack \cite{tu2023many}              & 0.014          & 0.005         & 0.208          & 0.165          & 0.098          & 0.002          & 0.008          \\
                             & MF-it \cite{zhao2023evaluating}                   & 0.101          & 0.042         & 0.279          & 0.456          & 0.22           & 0.017          & 0.062          \\
                             & MF-ii \cite{zhao2023evaluating}                    & 0.471          & 0.146         & 0.323          & 0.347          & 0.322          & 0.09           & 0.283          \\
                             & Attack-Bard \cite{dong2023robust}              & \textbf{0.934} & 0.146         & 0.487          & 0.278          & 0.461           & 0.098         &  0.386          \\
                             & CRAFT(ours)             & 0.662          & \textbf{0.531}  & \textbf{0.686} & \textbf{0.525} & \textbf{0.601} & \textbf{0.485} & \textbf{0.652}  \\ \hline
\multirow{6}{*}{Florence-2 \cite{xiao2024florence}}  & no attack               & 0.002          & 0.001         & 0.005          & 0.009          & 0.004          & 0              & 0             \\
                             & Mix.Attack \cite{tu2023many}              & 0.126          & 0.073         & 0.098          & 0.547          & 0.211          & 0.067          & 0.083         \\
                             & MF-it \cite{zhao2023evaluating}                   & 0.32           & 0.184         & 0.277          & 0.387          & 0.292          & 0.139          & 0.205          \\
                             & MF-ii \cite{zhao2023evaluating}                    & 0.571          & 0.332         & 0.501          & 0.462          & 0.466          & 0.264          & 0.338         \\
                             & Attack-Bard \cite{dong2023robust}             & \textbf{0.935} & 0.241         & 0.346          & 0.296          & 0.308          & 0.213          & 0.31          \\
                             & CRAFT(ours)             & 0.765          & \textbf{0.565}& \textbf{0.849} & \textbf{0.649} & \textbf{0.707} & \textbf{0.471} & \textbf{0.609} \\ \hline

\end{tabular}
\caption{Comparison of attack performance across four tasks (IC: Image Captioning, OD: Object Detection, RC: Region Categorization, OL: Object Localization) and evaluation metrics (avg: Average Success Rate, CTSR-4: Cross-Task Success Rate on all 4 tasks, CTSR-3: Cross-Task Success Rate on at least 3 tasks). Bold values indicate the best performance for each column.}
\label{main}
\end{table*}

\subsection{Cross-task Transferability of Task-specific Attacks}
\label{ts}
To further analyze the transferability of task-specific attacks compared to our cross-task approach, we implement Training Loss Minimization (TLM) attacks individually optimized for each of the four tasks. The TLM approach directly minimizes the task-specific training loss:
\begin{equation}
   \delta = \arg\min_{\|\delta\|_\infty \leq \epsilon} \mathcal{L}_{task}(\mathcal{M}(I+\delta, T_i), y_{tgt})
\end{equation}
where $\mathcal{L}_{task}$ is the task-specific loss function and $y_{tgt}$ is the target output.
Table~\ref{TLM-ctsr-4} presents the transferability comparison between task-specific TLM attacks and our CRAFT method. While each TLM attack achieves the highest success rate on its corresponding task (highlighted in bold), they show limited transferability to other tasks. CRAFT, despite not achieving the highest success rate on individual tasks, delivers substantially better cross-task performance with superior CTSR-4 and CTSR-3 scores. Interestingly, attacks optimized for Region Categorization (TLM-RC) demonstrate better transferability than those optimized for other tasks. This suggests that the Region Categorization task, which focuses on object identity within specific regions, shares more feature representations with other tasks. Our CRAFT method achieves its highest individual task performance on this task, which likely contributes to its superior cross-task transferability.

\begin{table}[ht]
\footnotesize
\centering
\begin{tabular}{cccccc}
\hline
\multirow{2}{*}{Method} & \multicolumn{4}{c}{Tasks}                                         & {Evaluate Metrics}   \\ \cline{2-6} 
                        & IC             & OD             & RC             & OL               & CTSR-4                 \\ \hline
TLM-IC                  & \textbf{0.935} & 0.241          & 0.346          & 0.296          & 0.213                   \\
TLM-OD                  & 0.451          & \textbf{0.827} & 0.683          & 0.518           & 0.316                   \\
TLM-RC                  & 0.523          & 0.548          & 0.703          & 0.724           & 0.42                   \\
TLM-OL                  & 0.244          & 0.257          & 0.277          & \textbf{0.736}  & 0.153                   \\
CRAFT(ours)             & 0.765          & 0.565          & \textbf{0.849} & 0.649          & \textbf{0.471}  \\ \hline
\end{tabular}

\caption{Performance comparison of CTSR-4 between task-specific Training Loss Minimization (TLM) attacks and our CRAFT method. TLM-IC, TLM-OD, TLM-RC, and TLM-OL represent attacks optimized specifically for Image Captioning, Object Detection, Region Categorization, and Object Location tasks, respectively. Bold values indicate the best performance for each column.}
\label{TLM-ctsr-4}
\end{table}

\begin{figure}[ht]
\centering
\includegraphics[width=\linewidth]{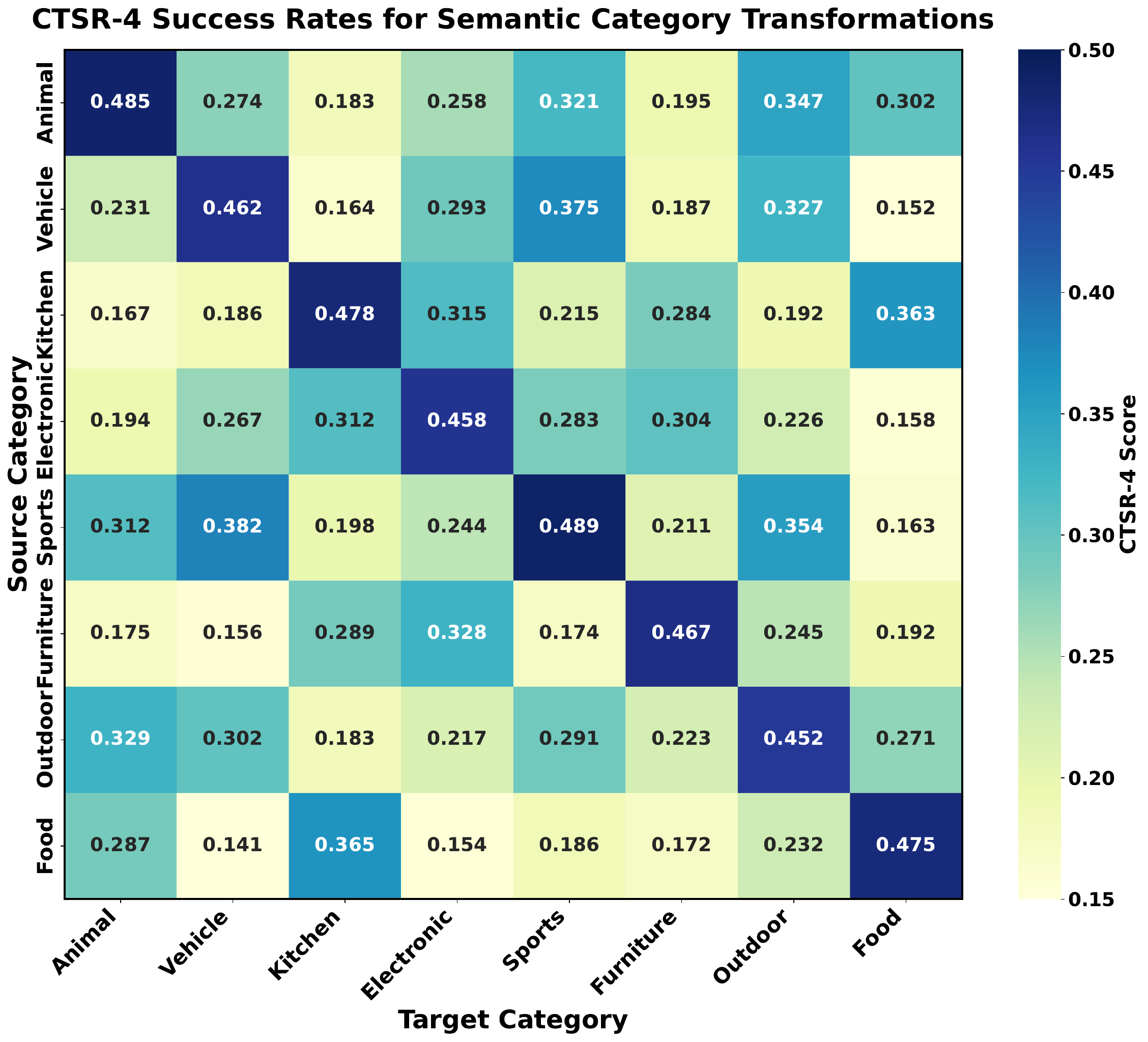}
\caption{Heatmap visualization of CTSR-4 success rates for semantic category transformations using CRAFT on Florence-2.}
\label{fig:semantic_heatmap}
\end{figure}

\subsection{Semantic Category Transformation Analysis}
\label{cross_cat}
To provide deeper insights into cross-task attack performance, we analyze the success rates across different semantic category transformations. Figure~\ref{fig:semantic_heatmap} presents a heatmap showing the CTSR-4 scores for transformations between semantic categories using our CRAFT method on Florence-2. The results reveal significant variation in transformation difficulty across different category pairs. Transformations within the same semantic category (e.g., Animal→Animal, Vehicle→Vehicle) generally achieve higher success rates, supporting the intuition that objects with similar structures and contexts are easier to manipulate across tasks. In contrast, cross-category transformations between dissimilar objects (e.g., Furniture→Electronic, Animal→Kitchen) prove substantially more challenging. Interestingly, certain cross-category transformations show unexpectedly high success rates, such as Vehicle→Sports and Animal→Outdoor. These exceptions may be attributed to contextual relationships (e.g., sports objects often appear near vehicles) or visual similarities that facilitate consistent manipulation across tasks despite category differences.These findings suggest that semantic similarity plays a significant role in cross-task target-change attack success.

\subsection{Ablation Studies and Analysis}
\label{ablation}
\begin{figure*}[ht]
\centering
\includegraphics[width=\linewidth]{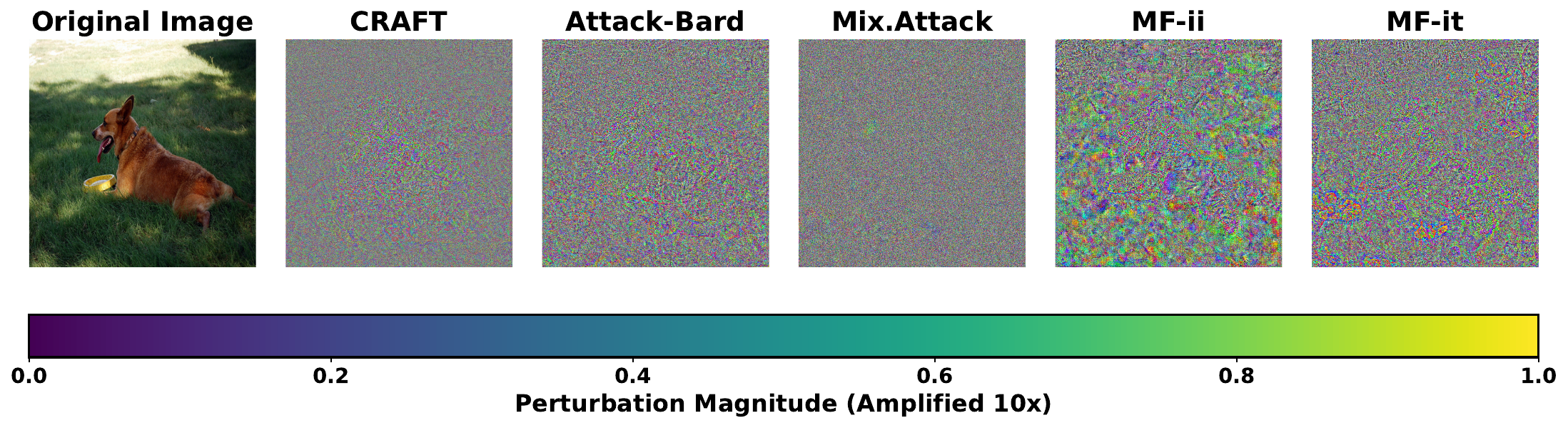}
\caption{Visualization of adversarial perturbations (amplified 10x for visibility) generated by different attack methods. CRAFT produces perturbations more concentrated around the target object region, while other methods generate more uniformly distributed noise patterns across the image.}
\label{fig:pv}
\end{figure*}

We conducted ablation studies on the Florence-2 model to assess our CRAFT method's key components, as shown in Table~\ref{tab:ablation}. The results indicate that Region Token Localization (RTL) markedly boosts attack performance, surpassing global image features regardless of the negative text strategy. As illustrated in Figure~\ref{fig:pv}, CRAFT's perturbations are concentrated on the target object, unlike other methods that generate diffuse noise. This precise manipulation of relevant features enhances cross-task deception while minimizing image alterations.
Regarding negative text selection, employing multiple negative texts (all non-target categories) with RTL generally outperforms using a single negative text (the source category) or none. This supports our theory that focusing perturbations on semantic regions while steering them clear of all incorrect categories improves cross-task transferability.

\begin{table}[htb]
\small
\centering
\begin{tabular}{ccccc}
\hline
\multirow{2}{*}{Metric} & \multirow{2}{*}{Features} & \multicolumn{3}{c}{Negative texts}                                                    \\ \cline{3-5} 
                        &                           & None & Source only & All others \\ \hline
\multirow{2}{*}{CTSR-4} & w/o RTL                   & 0.17  & 0.257       & 0.15       \\
                        & w/ RTL                    & 0.46  & 0.328       & 0.471      \\ \hline
\multirow{2}{*}{CTSR-3} & w/o RTL                   & 0.244 & 0.354       & 0.223      \\
                        & w/ RTL                    & 0.586 & 0.465       & 0.609      \\ \hline
\end{tabular}
\caption{Ablation study on Florence-2 model showing CTSR-4 and CTSR-3 performance with different combinations of image feature approaches (with or without Region Token Localization) and negative text selection strategies (None: no negative texts; Source only: only source category as negative; All others: all categories except target as negatives).}
\label{tab:ablation}
\end{table}

\begin{figure}[htb]
\centering
\includegraphics[width=\linewidth]{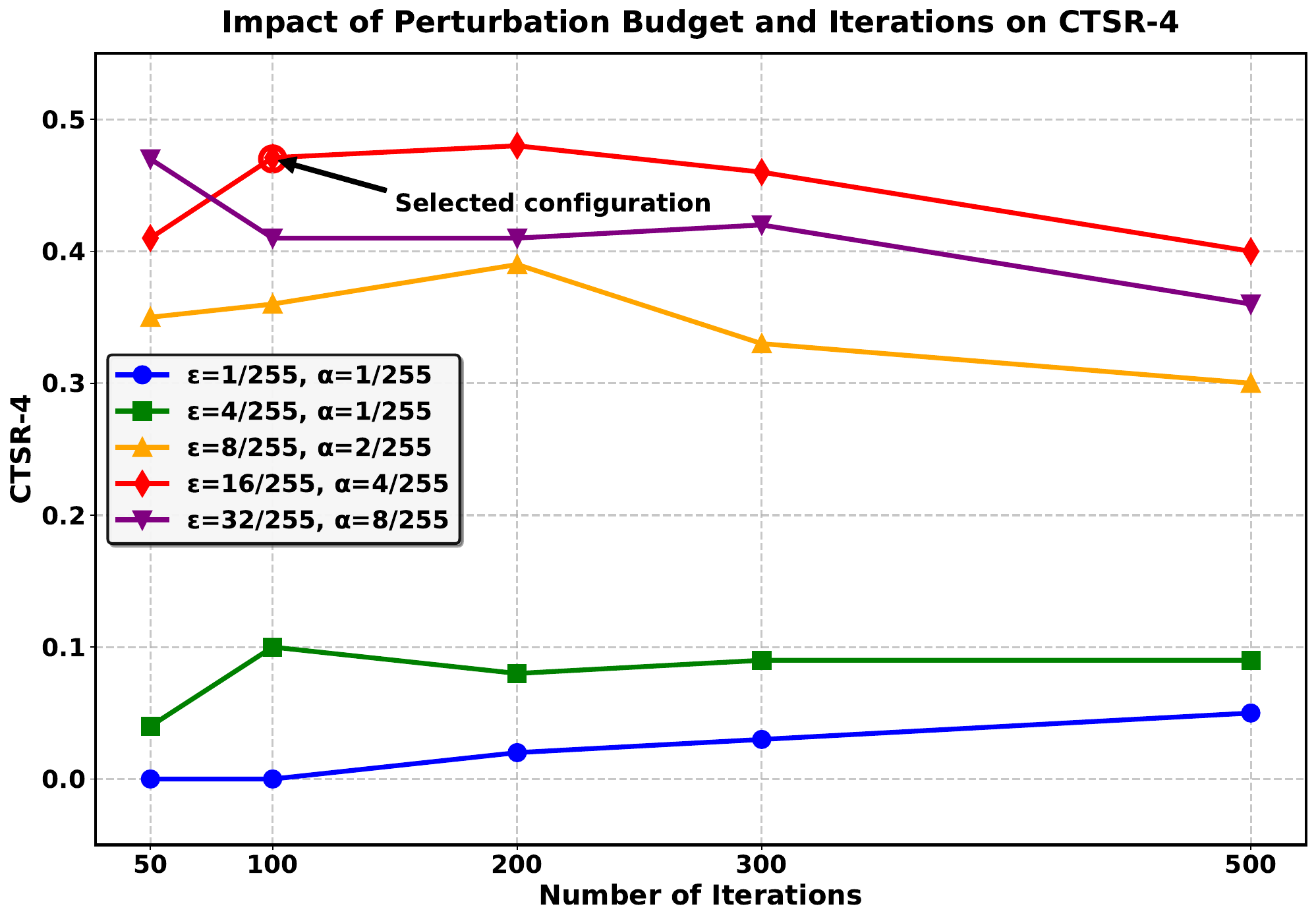}
\caption{Impact of perturbation budget ($\epsilon$) and iteration count on CTSR-4 performance with Florence-2 model. Each line represents a different $\epsilon$ value with corresponding step size $\alpha$.}
\label{fig:epsilon_iter_ctsr4}
\end{figure}

Figure~\ref{fig:epsilon_iter_ctsr4} illustrates the effect of perturbation budget ($\epsilon$) and iteration count on CTSR-4 performance. We observe that $\epsilon=16/255$ offers a good balance between effectiveness and visual imperceptibility, while 100 iterations strikes an optimal balance between computational efficiency and attack performance.
Interestingly, we observe that with extremely high iteration counts, cross-task attack performance occasionally decreases, likely due to overfitting to specific tasks at the expense of unified feature manipulation that benefits cross-task transferability.

\section{Conclusion}

This paper studies cross-task adversarial attacks on unified VLMs. We introduce CrossVLAD, a benchmark with new metrics to evaluate this threat, and propose CRAFT, a method that manipulates region-based, token-aligned features. Experiments show CRAFT's superior performance in cross-task attacks, revealing a key vulnerability in unified VLMs and underscoring the need for robust defenses against such deception.

\textbf{Acknowledgments.}  This work was supported by National Natural Science Fund of China (62473286).
{
    \small
    \bibliographystyle{ieeenat_fullname}
    \bibliography{main}
}

\clearpage
\setcounter{page}{1}
\maketitlesupplementary

\section{Dataset Construction Details}
\label{appendix:dataset}
\subsection{Comparison with Other Adversarial Attack Benchmarks}
Table \ref{tab:benchmark_comparison} compares CrossVLAD with existing adversarial attack benchmarks for vision-language models. While some prior works have considered unified VLMs or multiple tasks, CrossVLAD uniquely combines comprehensive multi-task coverage with explicit cross-task evaluation metrics. Additionally, our benchmark contains a substantial dataset size of 3,000 carefully curated samples, enabling statistically robust evaluation.

\begin{table}[h]
\small
\centering
\caption{Comparison with existing adversarial attack benchmarks for vision-language models.}
\label{tab:benchmark_comparison}
\resizebox{\linewidth}{!}{
\begin{tabular}{lccccc}

\toprule
Benchmark & Unified VLMs &  Multi-tasks & Cross-task Evaluation & Attack Type & Dataset Size \\
\midrule
CrossVLAD (Ours) & \checkmark  & \checkmark  & \checkmark & target & 3k \\
vllm-safety-bench\cite{tu2023many} & \checkmark &  \xmark & \xmark  & untarget & 2k \\
ROZ-benchmark\cite{wang2024benchmarking} & \xmark  & \xmark & \xmark & untarget &  \xmark \\
MultiTrust \cite{zhang2024benchmarking} & \checkmark & \xmark & \xmark & both & 200 \\
\bottomrule
\end{tabular}
}
\end{table}

\subsection{Selection Criteria}
The CrossVLAD benchmark was constructed using following filtering approach:

\begin{itemize}
    \item Excluded images containing potential target categories to avoid pre-existing confusion
    \item Verified source categories existed in the image
    \item Limited the maximum number of object instances per image to 5
    \item Ensured category uniqueness within each image to avoid ambiguity
    \item Object size constraints: Selected objects occupying between 10\% and 50\% of the image area
    \item Caption verification: Required objects to appear in at least 3 of the 5 MSCOCO captions
\end{itemize}

\subsection{Annotation Process}
For each selected image, we:
\begin{itemize}
    \item Preserved original MSCOCO annotations (bounding boxes, category labels)
    \item Randomly selected one qualified source object per image
    \item Identified appropriate target category from our predefined change-pairs
    \item Used GPT-4 to generate target captions with the following prompt:
\end{itemize}

\begin{lstlisting}
You are given a picture with a primary object called "[SOURCE_CATEGORY]". 
Below are 5 captions that describe the image including this object:
1. [CAPTION_1]
2. [CAPTION_2]
...
5. [CAPTION_5]

Task: Imagine replacing the primary object "[SOURCE_CATEGORY]" with a new object "[TARGET_CATEGORY]". Create a caption describing the scene with this replacement.
\end{lstlisting}

We implemented quality control by verifying that each generated caption: (1) explicitly mentioned the target category, (2) excluded the source category, and (3) maintained coherence with the original image context. Multiple generation attempts were made when necessary to ensure quality standards were met.

\subsection{Complete Change-Pair List}
The complete list of 79 change-pairs used in CrossVLAD is provided in table~\ref{pairs}

\begin{table*}[h]
\centering
\caption{Complete source-target object change pairs in CrossVLAD.}
\label{pairs}
\begin{tabular}{lll}
\toprule
\textbf{Category} & \textbf{Number of Pairs} & \textbf{Source $\rightarrow$ Target Examples} \\
\midrule
Vehicle & 8 pairs & bicycle $\rightarrow$ motorcycle, motorcycle $\rightarrow$ bicycle, car $\rightarrow$ bus, \\
& & bus $\rightarrow$ truck, train $\rightarrow$ airplane, truck $\rightarrow$ car, airplane $\rightarrow$ bus, boat $\rightarrow$ train \\
\midrule
Outdoor & 5 pairs & traffic light $\rightarrow$ stop sign, fire hydrant $\rightarrow$ stop sign, stop sign $\rightarrow$ traffic light, \\
& & parking meter $\rightarrow$ bench, bench $\rightarrow$ parking meter \\
\midrule
Animal & 10 pairs & bird $\rightarrow$ cat, cat $\rightarrow$ dog, dog $\rightarrow$ cat, horse $\rightarrow$ sheep, sheep $\rightarrow$ cow, \\
& & cow $\rightarrow$ horse, elephant $\rightarrow$ bear, bear $\rightarrow$ elephant, zebra $\rightarrow$ giraffe, \\
& & giraffe $\rightarrow$ zebra \\
\midrule
Accessory & 5 pairs & backpack $\rightarrow$ handbag, umbrella $\rightarrow$ handbag, handbag $\rightarrow$ suitcase, \\
& & tie $\rightarrow$ handbag, suitcase $\rightarrow$ backpack \\
\midrule
Sports & 10 pairs & frisbee $\rightarrow$ sports ball, skis $\rightarrow$ snowboard, snowboard $\rightarrow$ skateboard, \\
& & sports ball $\rightarrow$ kite, kite $\rightarrow$ baseball bat, baseball bat $\rightarrow$ baseball glove, \\
& & baseball glove $\rightarrow$ tennis racket, skateboard $\rightarrow$ surfboard, surfboard $\rightarrow$ skis, \\
& & tennis racket $\rightarrow$ frisbee \\
\midrule
Kitchen & 7 pairs & bottle $\rightarrow$ wine glass, wine glass $\rightarrow$ cup, cup $\rightarrow$ fork, fork $\rightarrow$ knife, \\
& & knife $\rightarrow$ spoon, spoon $\rightarrow$ bowl, bowl $\rightarrow$ bottle \\
\midrule
Food & 10 pairs & banana $\rightarrow$ apple, apple $\rightarrow$ orange, sandwich $\rightarrow$ hot dog, orange $\rightarrow$ banana, \\
& & broccoli $\rightarrow$ carrot, carrot $\rightarrow$ hot dog, hot dog $\rightarrow$ pizza, pizza $\rightarrow$ donut, \\
& & donut $\rightarrow$ cake, cake $\rightarrow$ apple \\
\midrule
Furniture & 6 pairs & chair $\rightarrow$ couch, couch $\rightarrow$ potted plant, potted plant $\rightarrow$ bed, \\
& & bed $\rightarrow$ dining table, dining table $\rightarrow$ toilet, toilet $\rightarrow$ chair \\
\midrule
Electronic & 6 pairs & tv $\rightarrow$ laptop, laptop $\rightarrow$ mouse, mouse $\rightarrow$ remote, remote $\rightarrow$ keyboard, \\
& & keyboard $\rightarrow$ cell phone, cell phone $\rightarrow$ tv \\
\midrule
Appliance & 5 pairs & microwave $\rightarrow$ oven, oven $\rightarrow$ toaster, toaster $\rightarrow$ sink, \\
& & sink $\rightarrow$ refrigerator, refrigerator $\rightarrow$ microwave \\
\midrule
Indoor & 7 pairs & book $\rightarrow$ clock, clock $\rightarrow$ vase, vase $\rightarrow$ scissors, scissors $\rightarrow$ teddy bear, \\
& & teddy bear $\rightarrow$ hair drier, hair drier $\rightarrow$ toothbrush, toothbrush $\rightarrow$ book \\
\bottomrule
\end{tabular}
\end{table*}

\section{Pseudocode}
\label{appendix:algorithm}
Algorithm~\ref{alg:craft} presents the detailed procedure of our CRAFT method. 
The procedure begins by initializing the adversarial example and locating the token indices corresponding to the source object region. In each iteration, we extract image features from the current adversarial example and isolate the features of the target region. We then obtain text embeddings for both positive (target) and negative (source and other) categories. The contrastive loss is computed to align the region features with the target category while pushing them away from negative categories. Finally, we update the adversarial example using the PGD algorithm with the computed gradient.

\begin{algorithm}
\caption{CRAFT: Cross-task Region-based Attack Framework with Token-alignment}
\label{alg:craft}
\begin{algorithmic}[1]
\REQUIRE Input image $I$, source object bounding box $b_s$, source category $c_s$, target category $c_t$, perturbation budget $\epsilon$, iteration number $T$, step size $\alpha$
\ENSURE Adversarial example $I_{adv}$
\STATE $I_{adv} \leftarrow I$ \COMMENT{Initialize adversarial example}
\STATE $\mathcal{R} \leftarrow \text{RegionTokenLocalization}(b_s)$ \COMMENT{Localize tokens corresponding to source object}
\FOR{$t = 1$ to $T$}
   \STATE $F_I \leftarrow \text{ImageEncoder}(I_{adv})$ \COMMENT{Extract image feature tokens}
   \STATE $F_R \leftarrow F_I[\mathcal{R}]$ \COMMENT{Extract region feature tokens}
   
   \STATE $E_{pos} \leftarrow \text{TextEncoder}(c_t)$ \COMMENT{Encode target category}
   \STATE $E_{neg} \leftarrow \text{TextEncoder}(c_s, \text{other categories})$ \COMMENT{Encode negative categories}
   
   \STATE $\mathcal{L} \leftarrow \text{ContrastiveLoss}(F_R, E_{pos}, E_{neg})$ \COMMENT{Compute alignment loss}
   
   \STATE $g \leftarrow \nabla_{I_{adv}}\mathcal{L}$ \COMMENT{Compute gradient}
   \STATE $I_{adv} \leftarrow \text{Clip}(I_{adv} + \alpha \cdot \text{sign}(g), I-\epsilon, I+\epsilon)$ \COMMENT{Update with PGD}
\ENDFOR
\RETURN $I_{adv}$
\end{algorithmic}
\end{algorithm}

\section{Experimental Details}
\label{exp_detail}
\subsection{Implementation Details of Compared Methods}

We provide detailed implementation information for all compared methods to ensure reproducibility and fair comparison.

\paragraph{Attack-Bard}
We adopt the text description attack from Attack-Bard \cite{dong2023robust}, which maximizes the log-likelihood of predicting a target sentence. For our evaluation, we use GPT-generated captions that include the target object category while excluding the source object category. The attack is formulated as:

\begin{equation}
\max_{x} \sum_{i=1}^{N} \sum_{t=1}^{L} \log p_{\theta_i}(y_t|x, p, y_{<t}), \quad \text{s.t.} \quad \|x - x_{nat}\|_{\infty} \leq \epsilon
\end{equation}

where $y_t$ represents tokens in the target caption, $p$ is the prompt, and $\theta_i$ denotes model parameters. We optimize this objective using the PGD algorithm with the same hyperparameters as our primary method.

\paragraph{Mix.Attack}
Since the original Mix.Attack \cite{tu2023many} was designed for untargeted attacks, we modified it for our targeted setting. Our adaptation aligns the adversarial image with the target text while pushing it away from the original descriptions. Specifically, we use three text references: two from the original MSCOCO caption annotations and one from our generated target caption labels. The optimization objective encourages similarity between the image and target caption representations while reducing similarity with the original captions. All image and text features are extracted using the attacked model's own encoders to ensure alignment with the model's internal representations.

\paragraph{MF-it}
For MF-it \cite{zhao2023evaluating}, we directly compute the similarity between image features and target caption text features, then minimize this similarity through adversarial optimization. This approach attempts to align the perturbed image's feature representation with the textual representation of the target category. The optimization is performed using PGD with the same constraints as our main experiments.

\paragraph{MF-ii}
For MF-ii\cite{zhao2023evaluating}, we first generate a target image using Stable Diffusion \cite{rombach2022high} with our target caption label as the prompt. We then minimize the feature distance between the adversarial image and this generated target image. This approach attempts to make the adversarial image perceptually similar to an image of the target category while maintaining the $\ell_{\infty}$ perturbation constraint.

\subsection{Additional Experimental Results}

\begin{table}[ht]
\footnotesize
\centering
\begin{tabular}{cccccc}
\hline
\multirow{2}{*}{Method} & \multicolumn{4}{c}{Tasks}                                         & {Evaluate Metrics}   \\ \cline{2-6} 
                        & IC             & OD             & RC             & OL               & CTSR-3                 \\ \hline
TLM-IC                  & \textbf{0.935} & 0.241          & 0.346          & 0.296          & 0.305                   \\
TLM-OD                  & 0.451          & \textbf{0.827} & 0.683          & 0.518           & 0.54                   \\
TLM-RC                  & 0.523          & 0.548          & 0.703          & 0.724           & 0.568                   \\
TLM-OL                  & 0.244          & 0.257          & 0.277          & \textbf{0.736}  & 0.252                   \\
CRAFT(ours)             & 0.765          & 0.565          & \textbf{0.849} & 0.649          & \textbf{0.609}  \\ \hline
\end{tabular}

\caption{Performance comparison of CTSR-3 between task-specific Training Loss Minimization (TLM) attacks and our CRAFT method. TLM-IC, TLM-OD, TLM-RC, and TLM-OL represent attacks optimized specifically for Image Captioning, Object Detection, Region Categorization, and Object Location tasks, respectively. Bold values indicate the best performance for each column.}
\label{TLM-ctsr-3}
\end{table}

\subsubsection{Cross-task Transferability of Task-specific Methods Results with CTSR-3}
Table~\ref{TLM-ctsr-3} presents the transferability comparison between task-specific TLM attacks and our CRAFT method using the CTSR-3 metric. These results complement those in Section~\ref{ts} by showing performance when success on at least three tasks is required rather than all four. While the overall trends remain consistent with the CTSR-4 results, the higher CTSR-3 values provide additional insights into partial transferability. Notably, the gap between task-specific attacks and CRAFT is smaller under this more relaxed evaluation criterion, though CRAFT still maintains its advantage, particularly for tasks with spatial components.

\subsubsection{Ablation Study with CTSR-3}
\begin{figure}[h]
\centering
\includegraphics[width=0.9\linewidth]{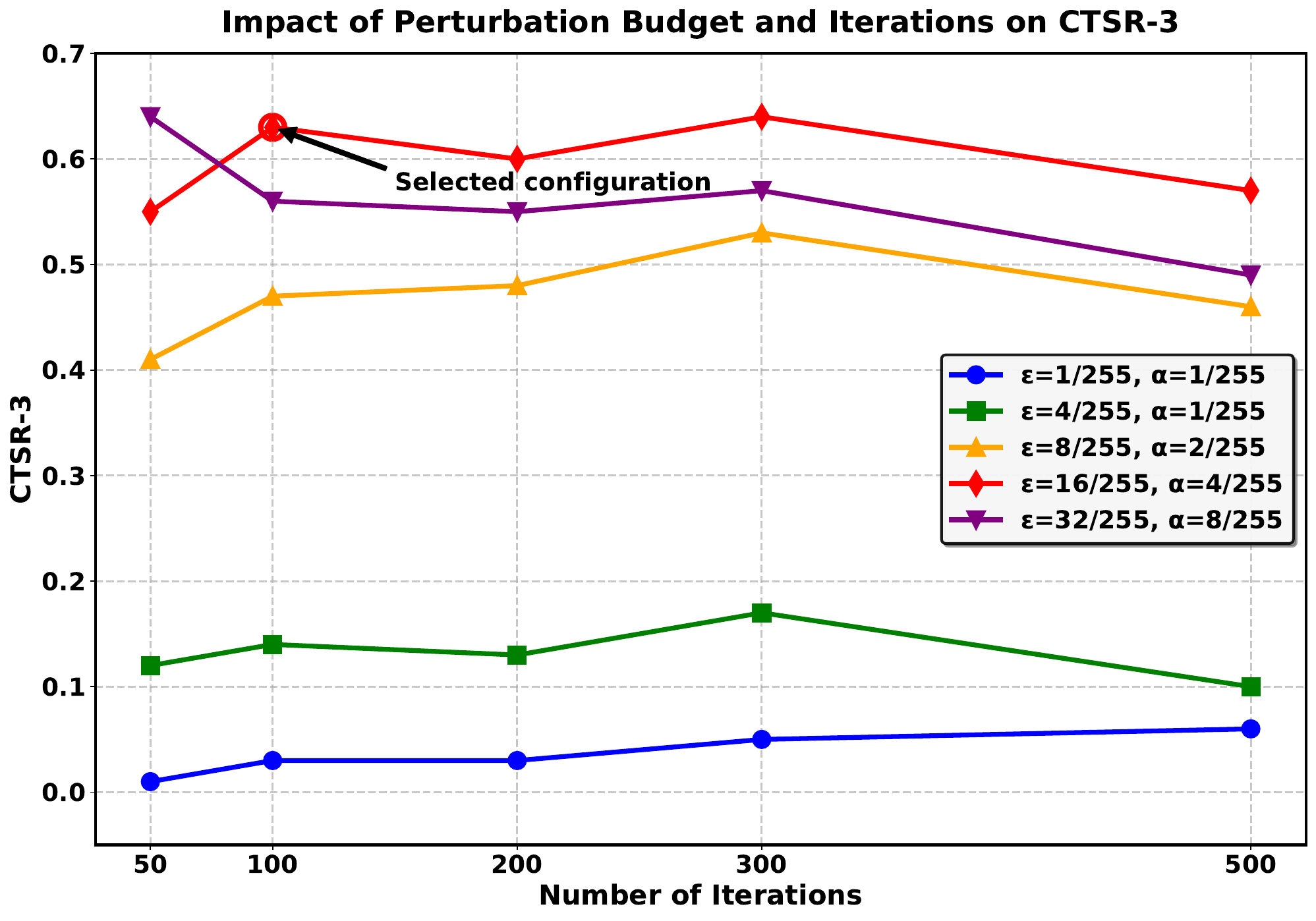}
\caption{Impact of perturbation budget ($\epsilon$) and iteration count on CTSR-3 performance with Florence-2 model. Each line represents a different $\epsilon$ value with corresponding step size $\alpha$.}
\label{fig:epsilon_iter_ctsr3}
\end{figure}

Figure~\ref{fig:epsilon_iter_ctsr3} shows the impact of perturbation budget ($\epsilon$) and iteration count on CTSR-3 performance. The trends broadly mirror those observed for CTSR-4 in Section~\ref{ablation}, but with higher overall success rates as expected from the more lenient evaluation criterion. The CTSR-3 results further support our choice of $\epsilon=16/255$ and 100 iterations as the optimal configuration, offering a favorable balance between attack success rate and computational efficiency. Interestingly, the performance plateau and potential decrease with very high iteration counts is less pronounced for CTSR-3, suggesting that overfitting to specific tasks is less problematic when success on only three out of four tasks is required.

\subsection{Additional Visualization Examples}
\begin{figure*}[ht]
\centering
\includegraphics[width=\linewidth]{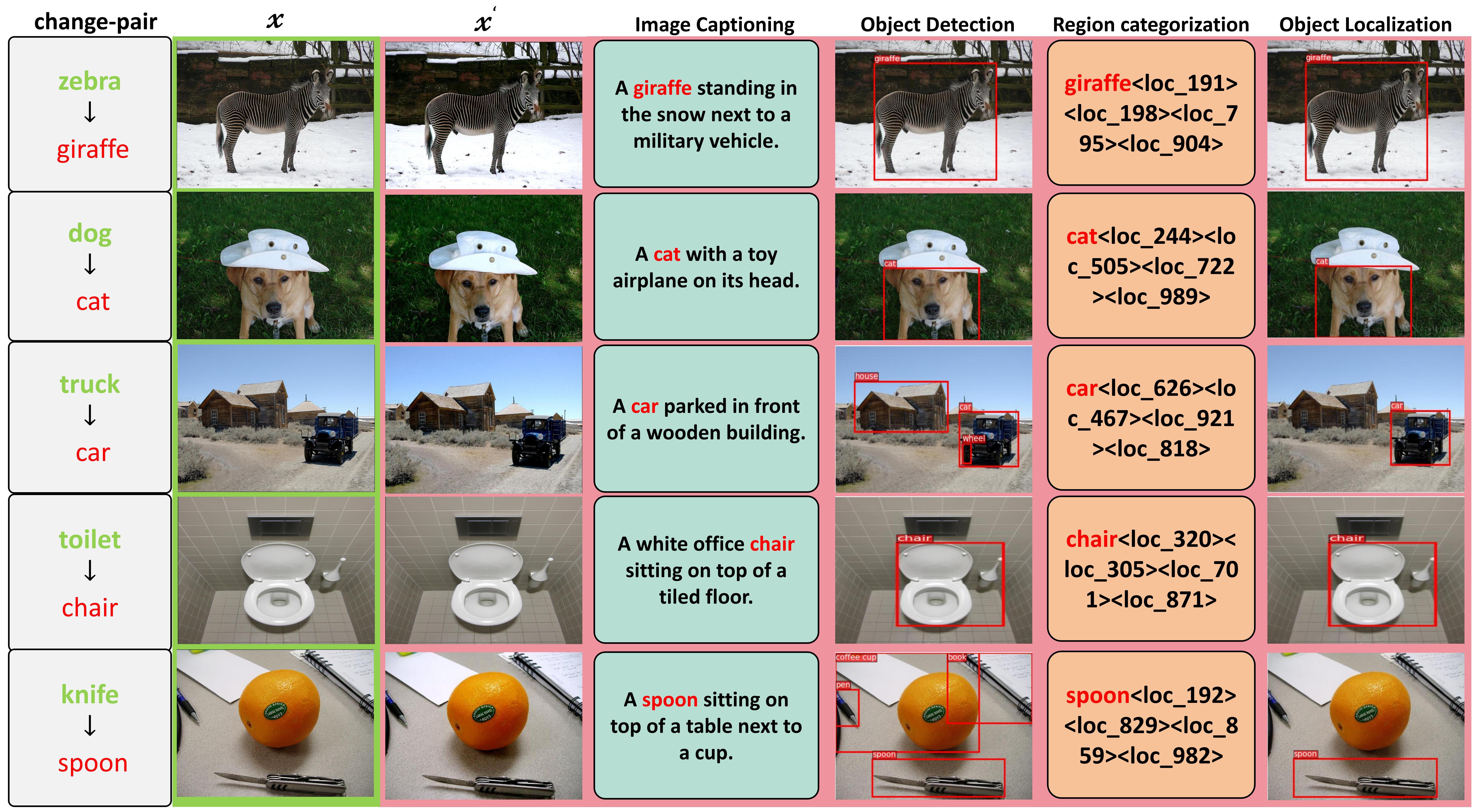}
\caption{Qualitative examples of CRAFT attack on Florence-2 model across four vision tasks: object detection, object localization, image captioning, and region categorization, with $\varepsilon=16/255$.}
\label{appen_demo}
\end{figure*}

\begin{figure*}[ht]
\centering
\includegraphics[width=\linewidth]{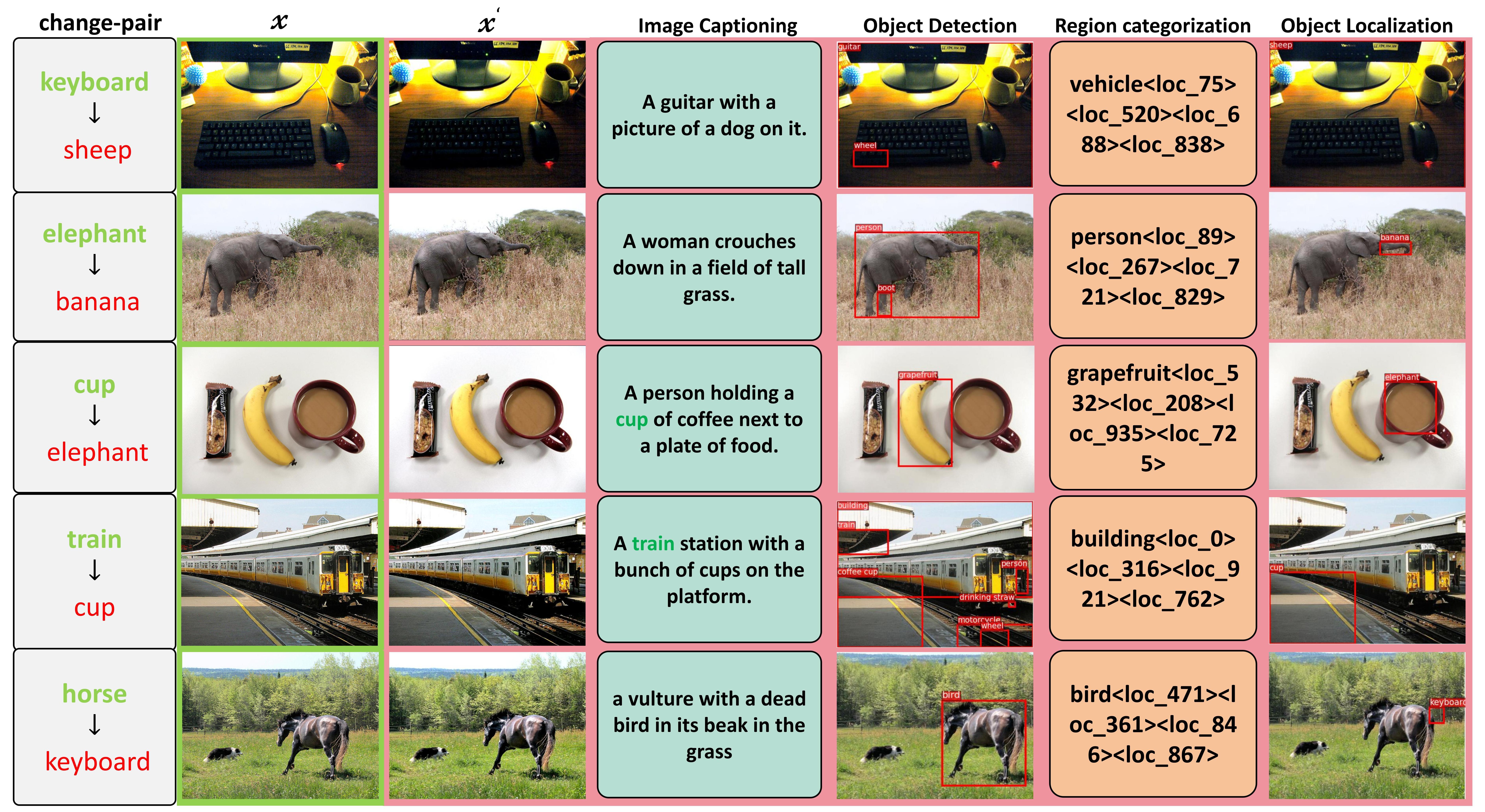}
\caption{Failed examples of CRAFT attack on Florence-2 model across four vision tasks: object detection, object localization, image captioning, and region categorization, when using cross-category substitutions, with $\varepsilon=16/255$.}
\label{fail}
\end{figure*}

Figure~\ref{appen_demo} illustrates additional successful examples, while Figure~\ref{fail} complements Section~\ref{cross_cat} by demonstrating that target replacement often becomes unfeasible when modifications occur across categories. However, it can be observed that in most instances, these cross-category alterations also disrupt the original semantic content of the image, enabling untargeted attacks.

\subsection{Evaluation on commercial VLMs}

While our primary focus is on white-box models where access to model parameters is necessary, we also conducted a preliminary evaluation on the black-box transferability of our adversarial attacks to commercial Vision Language Models (VLMs). To this end, we investigate whether adversarial examples generated for an open-source model can successfully deceive a closed-source, commercial model.

\begin{figure}[h]
\vspace{-10pt}
\centering
\includegraphics[width=\linewidth]{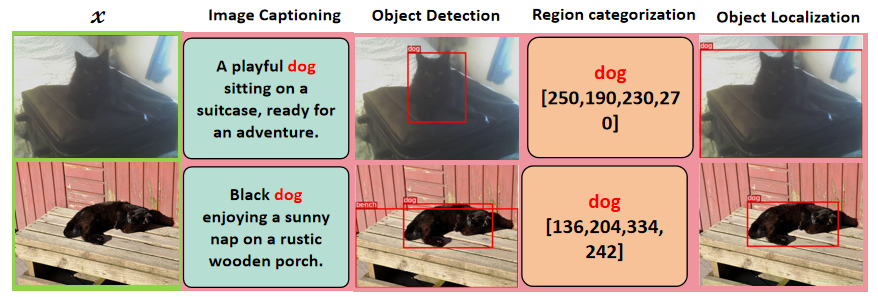}
\caption{Successful cat-to-dog attacks transferring to GPT-4V.}
\label{fig:gpt4v}
\vspace{-15pt}
\end{figure}

Specifically, we generated adversarial examples using Florence-2, a publicly available model, with the objective of causing a cat-to-dog misclassification. These generated images were then presented to GPT-4V, a prominent commercial VLM, to assess the transferability of the attack. As illustrated in Figure \ref{fig:gpt4v}, the adversarial examples crafted on Florence-2 were effective in misleading GPT-4V, which consequently misidentified the cats in the images as dogs. This successful transfer demonstrates the practical relevance of our attack methodology in a black-box setting, even though it was conducted on a limited scale.

\subsection{Comparison with Object Detection Attack Methods}
To further contextualize the performance of our method, we conducted a comparative analysis against established attack methodologies originally designed for object detection.

The results of this comparison on the Florence-2 model are presented in Table \ref{od_baseline}. The findings indicate that while the adapted baseline methods  achieve strong performance on the object detection (OD) task itself, their effectiveness diminishes significantly when transferred to other vision-language tasks such as Image Captioning (IC), Region Categorization (RC), and Object Localization (OL).
In contrast, our method, CRAFT, demonstrates superior cross-task effectiveness. Although it shows slightly lower performance on the OD task, it significantly outperforms both baselines across all other evaluated tasks and achieves the highest Cross-Task Success Rates (CTSR-4 and CTSR-3).

\begin{table}[h]
\vspace{-5pt}
\centering
\scriptsize
\begin{tabular}{lcccccc}
\toprule
\textbf{Method} & \textbf{IC↑} & \textbf{OD↑} & \textbf{RC↑} & \textbf{OL↑} & \textbf{CTSR-4↑} & \textbf{CTSR-3↑} \\
\midrule
\cite{cai2022context}  & 0.48 & \textbf{0.75} & 0.68 & 0.52 & 0.38 & 0.52  \\
\cite{nezami2021pick} & 0.45 & 0.67 & 0.64 & 0.50 & 0.32 & 0.48 \\
CRAFT (ours) & \textbf{0.77} & 0.57 & \textbf{0.85} & \textbf{0.65} & \textbf{0.47} & \textbf{0.61}  \\
\bottomrule
\end{tabular}
\caption{Comparison with object detection attack baselines on Florence-2. Our method demonstrates superior cross-task transferability.}
\label{od_baseline}
\vspace{-15pt}
\end{table}

\subsection{Effect of Bounding Box Source}
In our primary experiments, we assume access to ground-truth bounding boxes to define the target regions for our attacks. To assess the practical applicability of our method in scenarios where such ground-truth data is unavailable, we conducted an ablation study to evaluate the impact of using bounding boxes generated by a state-of-the-art object detector.

For this analysis, we replaced the ground-truth bounding boxes with boxes detected by YOLOv10. We then performed the same attack procedure on the Florence-2 model. The comparative results are detailed in Table \ref{table:od}. The data shows that the performance difference between using YOLOv10-detected boxes and ground-truth boxes is minimal across all evaluated tasks. This experiment demonstrates that our attack's effectiveness is not contingent on having perfect bounding box information and that comparable results can be achieved using high-quality, readily available object detectors.

\begin{table}[h]
\vspace{-5pt}
\centering
\scriptsize
\begin{tabular}{lcccccc}
\toprule
\textbf{Box Source} & \textbf{IC↑} & \textbf{OD↑} & \textbf{RC↑} & \textbf{OL↑} & \textbf{CTSR-4↑} & \textbf{CTSR-3↑} \\
\midrule
YOLOv10 & 0.75 & 0.53 & 0.81 & 0.64 & 0.45 & 0.60  \\
Ground-truth & \textbf{0.77} & \textbf{0.57} & \textbf{0.85} & \textbf{0.65} & \textbf{0.47} & \textbf{0.61}  \\
\bottomrule
\end{tabular}
\caption{Performance comparison using bounding boxes from a SOTA detector (YOLOv10) versus ground-truth boxes. The minimal difference validates our experimental assumption.}
\label{table:od}
\vspace{-15pt}
\end{table}

\end{document}